\pdfoutput=1

\documentclass[11pt]{article}

\usepackage[]{emnlp2021}

\usepackage{times}
\usepackage{latexsym}
\usepackage{url}

\usepackage[T1]{fontenc}

\usepackage[utf8]{inputenc}

\usepackage{microtype}

\title{{BERT}weet{FR} : Domain Adaptation of Pre-Trained Language Models for French Tweets}

\author{Yanzhu Guo \\ École Polytechnique, France \\Shanghai Jiao Tong University, China \\ \texttt{yanzhu.guo@polytechnique.edu} \\ \And
  Virgile Rennard \\ École Polytechnique, France \\ \texttt{virgile@rennard.org} \\
  \AND
  Christos Xypolopoulos \\ École Polytechnique, France \\ \texttt{christos.xypolopoulos@polytechnique.edu} \\
  \AND
  Michalis Vazirgiannis \\ École Polytechnique, France \\ \texttt{mvazirg@lix.polytechnique.fr} \\}

\begin{document}
\maketitle
\begin{abstract}
We introduce \textbf{BERTweetFR}, the first large-scale pre-trained language model for French tweets. Our model is initialized using the general-domain French language model CamemBERT \cite{martin2020camembert} which follows the base architecture of RoBERTa. Experiments show that BERTweetFR outperforms all previous general-domain French language models on two downstream Twitter NLP tasks of offensiveness identification and named entity recognition. The dataset used in the offensiveness detection task is first created and annotated by our team, filling in the gap of such analytic datasets in French. We make our model publicly available in the transformers library with the aim of promoting future research in analytic tasks for French tweets.
\end{abstract}

\section{Introduction}

Vector representations of words have given rise to the application of deep learning methods in NLP. Traditional pre-training approaches, such as word2vec \cite{mikolov2013efficient} and GloVe \cite{pennington2014glove} are static, learning a single representation for every word regardless of context. However, words are often polysemous with different meanings depending on the context. More recently, models are trained to integrate contextual meaning : the output word embeddings depend on the whole input sequence rather than only the word itself. While the idea was initially implemented with recurrent neural networks \cite{daisemi} \cite{ramachandran2017unsupervised}, recent models have predominantly been based on the transformers architecture \cite{vaswani2017attention}, with BERT \cite{devlin2019bert} and RoBERTa \cite{liu2019roberta} among the most popular. These contextualized word representation models have opened new doors for researchers as they can be applied to numerous downstream tasks by fine-tuning or prompting. In fact, large-scale pre-trained language models have become the go-to tool for building new NLP applications, saving researchers from the enormous amount of computational resource and data required for training model weights from scratch.

In the past few years, human society has become more digitally connected than ever before. People use social media to report the latest news, but also to express their opinions and feelings about real-world events. As one of the most popular micro-blogging platforms, Twitter has become a primary source for social media user-generated data \cite{ghani2019social}. However, tweets are more often written in informal language compared to the carefully edited texts that are published in traditional data sources such as Wikipedia and printed media. They have their own set of features such as the recurrent use of irregular or abbreviated words, the large quantity of spelling or grammatical mistakes, the employment of improper sentence structures and the occurrence of mixed languages \cite{farzindar2020natural}. This presents challenges to standard NLP methods when applied to Twitter data.

Domain-adaptive pre-training has been revealed to provide significant gains in helping models encode the complexity of specific textual domains \cite{gururangan2020don}. While efforts on domain adaptation of large-scale language models to Twitter language have  been made in English \cite{nguyen2020bertweet}, there has been no similar work in any other language. 

We start addressing this problem by releasing BERTweetFR, a pre-trained language model for French tweets together with a manually labeled dataset for offensiveness identification in French tweets. We evaluate our model on two downstream tasks : offensiveness identification and named entity recognition.

We compare the performance of our model to the best-performing transformer-based models pre-trained on general domain French texts, namely CamemBERT \cite{martin2020camembert}, FlauBERT \cite{leflaubert} and BARThez \cite{eddine2020barthez}. Experiments show that our model outperforms all three of them on both of the downstream tasks.

In order to facilitate future research on French tweets, we make our model publicly available on Huggingface's model hub \footnote{\url{https://huggingface.co/Yanzhu/bertweetfr-base}} as well as on our team's website dedicated to French linguistic resources \footnote{\url{http://master2-bigdata.polytechnique.fr/FrenchLinguisticResources/bertweetFr}}. We will also release our offensiveness identification dataset while respecting the limitations of Twitter's developper policy.

\section{Building the Model}
In this section, we describe the collection steps and pre-procesing pipeline for our pre-training dataset, present the model architecture and introduce the training objective and optimization setup.

\subsection{Pre-training Data}
We use a 16GB dataset of 226M deduplicated French tweets. The tweets are deduplicated using open-source tool runiq\footnote{\url{https://github.com/whitfin/runiq}}. In addition, we filter out tweets with fewer than 5 tokens assuming they do not contain substantial information. The average length of a tweet is 30 tokens.

\subsubsection{Data Collection}
\label{data}

Our final dataset for pre-training is an aggregation of three corpora from different sources. The aggregation of these corpora makes this dataset the largest one for French tweets up to this date. It is also beneficial to aggregate different corpora in order to cover tweets from different time periods with diverse topics and styles. 

The three sources we use are as follows :
\begin{itemize}
    \item We start by downloading tweets from the general Twitter Stream\footnote{\url{https://archive.org/details/ twitterstream}} grabbed by the Archive Team, containing of tweets streamed from January 2016 to December 2019. Selecting only the French tweets with Twitter's built-in feature, we obtain a corpus of $34M$ unique tweets.
    
    \item We also build a COVID-19 related corpus of French tweets relevant to the COVID-19 pandemic posted between September 2020 and April 2021. In this case, our filters are focused on tweets that include the hashtags “covid19” and “coronavirus”. Through Twitter’s public streaming API, we extracted tweets in French marked with either or both of the two above hashtags. This corpora consists of $19M$ unique tweets.
    
    \item Finally, we make use of a previous Twitter dataset constructed for socioeconomic analysis \cite{abitbol2018socioeconomic}. This corpus includes a collection of tweets in French between the years 2014 and 2018. We extract $173M$ unique tweets form this corpus.
    
\end{itemize}

\subsubsection{Data Pre-Processing}
We only implement minimal data cleaning before inputting the sequences into the tokenizer. Following \cite{nguyen2020bertweet}, we normalize the Tweets by converting user mentions and web/url links into special tokens @USER and HTTPURL.

For tokenization, we apply the CamemBERT tokenizer \cite{martin2020camembert}. The CamemBERT tokenizer segments input sequences into subword units using the SentencePiece \cite{kudo2018sentencepiece} algorithm. This algorithm is an extension of Byte-Pair encoding (BPE) \cite{shibata1999byte} and WordPiece \cite{kudo2018subword} that eliminates the pre-tokenization step, thus more generally applicable. The vocabulary size is $32k$ subword tokens.

\subsection{Model Architecture}

\label{architecture}

BERTweetFR is initialized from the base version of CamemBERT. We choose to further fine-tune this model instead of starting from scratch because domain-adaptive pre-training have been proven to give very satisfying results in numerous downstream tasks spanning across a wide range of domains \cite{gururangan2020don}. The choice to employ CamemBERT instead of other French language models is based on its overall best performance on downstream tasks experimented in previous works \cite{eddine2020barthez}.

CamemBERT applies the multi-layer bidirectional Transformer architecture. The base version uses the same architectures as the base version of BERT with 12 layers, 768 hidden dimensions and 12 attention heads, adding up to a total of $110M$ parameters. CamemBERT follows the same optimized pre-training approach as RoBERTa, the only difference is that it uses whole-word masking and the SentencePiece tokenization instead of WordPiece.

\begin{table*}
  \small
  \caption{Offensive Identification scores on the best model selected.}
  \centering
  \label{offensive}
  \begin{tabular}{c|llll}
    \toprule
     & CamemBERT & FlauBERT & BARThez & \textbf{BERTweetFR} \\
    \midrule
    
    Accuracy & 86.47 & 86.87 & 84.35 & \textbf{88.07}  \\

    F1 Score & 68.89 & 65.20 & 67.51 & \textbf{71.27}  \\
    
    \bottomrule
  \end{tabular}
\end{table*}

\begin{table*}
  \small
  \caption{NER score on the best model selected.}
  \centering
  \label{ner}
  \begin{tabular}{c|llll}
    \toprule
     & CamemBERT & FlauBERT  & \textbf{BERTweetFR} \\
    \midrule
    Accuracy & 94.78 & 94.73  & \textbf{94.99}  \\
    F1 Score & 61.01 & 60.57  & \textbf{62.77}  \\
    
    \bottomrule
  \end{tabular}
\end{table*}

\subsection{Training Objective}
Our model is trained on the Masked Language Modeling (MLM) task. With any given input sequence, $15\%$ of the tokens are chosen for possible replacement. Among the selected tokens, $80\%$ are further selected to be replaced by the special <MASK> token, $10\%$ remain unchanged and $10\%$ are replaced by random tokens. Finally, the model is trained to predict the tokens replaced by <MASK> using cross-entropy loss. Following RoBERTa, we do not fix the whole set of masked tokens during pre-processing but select them dynamically during the training process. The data is thus augmented when training for multiple epochs.

\subsection{Optimization Setup}

We employ the CamemBERT implementation in the transformers library \cite{wolf2020transformers}. The maximum sequence length is set to be 128, generating approximately $226M \times 30 / 128 \approx 53M$ sequence blocks. Following \cite{gururangan2020don}, we optimize the model using Adam \cite{DBLP:journals/corr/KingmaB14} with a batch size of 1280 across 8 V100 GPUs (32GB each) and a peak learning rate of 0.0001. We pre-train the model for 20 epochs in about 8 days with a total of $53M \times 20 / 1280 \approx 83K$ training steps.

\section{Downstream Task Datasets}
We evaluate the performance of BERTweetFR on two downstream Twitter NLP tasks. The datasets used in these downstream tasks are either constructed by ourselves or obtained from shared tasks in past conferences.

\subsection{Offensive Language Identification}

Along with the outbreak of COVID-19 came severe disruption in the French society. Effects include the public holding the government accountable for certain ways in which the pandemic was handled, as well as a rise of hateful sentiment towards the Asian community. In fact, ever since the explosion of COVID-19 on a global scale, unrest has led to an increase of violent incidents towards people of Asian descent.

In response to this phenomenon, we have created a human annotated dataset for general offensiveness detection in Tweets collected during the COVID-19 pandemic. Our dataset contains 5786 French tweets among which 1301 have been labeled as offensive. Offensiveness is not straightforward and can be subjective. In our labeling procedure, we consider a tweet as offensive in cases where personal attacks are detected. For example, "The chinese virus is tiring" would not be considered offensive, while "I hate the chinese for bringing us the chinese virus" would be. This is a binary sequence classification task. We randomly sample a 70/15/15 training/validation/test split with each class proportionally represented in each part of the split.

\subsection{Named Entity Recognition}
For the NER task, we take data from the CAp 2017 challenge \cite{lopez2017cap}. This challenge proposes a new benchmark for the problem of NER for tweets written in French. The tweets were collected using the publicly available Twitter API and annotated with 13 types of entities : person, musicArtist, organisation, geoLoc, product, transportLine, media, sportsTeam, event, tvShow, movie, facility and other. Overall, the dataset comprises 6685 annotated tweets split into two parts: a training set consisting of 3000 tweets and a test set with 3685 tweets. For compatibility with previous research, the data were released tokenized using the CoNLL format and the BIO encoding.

\section{Baselines}
As our BERTweetFR model is the first pre-trained language model for French tweets, we compare it with the following general-domain language models for French. 


\paragraph{CamemBERT}
As mentioned in \ref{architecture}, CaemBERT \cite{martin2020camembert} is the model from which we started our fine-tuning for domain adaptation. It therefore serves as a natural baseline. Its architecture is already introduced in \ref{architecture}.

\paragraph{FlauBERT}
FlauBERT \cite{leflaubert} is another transfomer-based model trained on a very large and heterogeneous French corpus. It basically follows the same architecture as CamemBERT and is shown to outperform CamemBERT on some of the downstream tasks.

\paragraph{BARThez}
BARThez \cite{eddine2020barthez} is the first French sequence-to-sequence model based on the base version of the BART architecture \cite{lewis2020bart}. It has 6 encoder and 6 decoder layers with 768 hidden dimensions and 12 attention heads in both the encoder and the decoder. It is shown to be competitive in comparison with CamemBERT and FlauBERT.

\section{Experiments and Results}

In this section, we first describe our fine-tuning approaches. The baseline models follow the same fine-tuning procedures as BERTweetFR. We then report our results and compare with the baselines. As a result, our model substantially outperforms all baselines.

\subsection{Offensive Language Identification}

The offensive language identification task is a supervised sequence classification task. Following \cite{devlin2019bert}, we append a linear prediction layer on top of the pooled output. 

For fine-tuning, we employ transformers library to train BERTweetFR on the training set for 15 epochs. We use AdamW \cite{loshchilov2018decoupled} with a fixed learning rate of 2.e-5 and a batch size of 32 following \cite{liu2019roberta}. We compute the classification accuracy and F1 score after each training epoch on the validation set, applying early stopping if their is no improvement after 3 consecutive epochs. We eventually select the model checkpoint with the highest F1 score to predict the final labels on the test set. Our results are listed in Table \ref{offensive}.

\subsection{Named Entity Recognition}

The NER task is a supervised token classification task. Following \cite{devlin2019bert}, we append a linear prediction layer on top of the last Transformer layer with regards to the first subword of each word token. 

For fine-tuning, we again employ transformers library to train for 30 epochs. We use AdamW \cite{loshchilov2018decoupled} with a fixed learning rate of 2.e-5 and a batch size of 32, adding in weight decay. We compute performance scores for each entity class as well the overall F1 Micro score. We eventually select the model checkpoint with the highest F1 Micro score to predict the final labels on the test set. Our results are listed in Table \ref{ner}. We do not compare with BARThez in this task because the original model is not implemented for sequence classification tasks.

\section{Conclusion}
In this work, we investigated the effectiveness of applying domain adaptation to the Twitter domain for large-scale pre-trained French language models. We demonstrate the value of our model showing that it outperforms all previous general-domain French language models on two downstream Twitter NLP tasks of offensiveness identification and named entity recognition.

Our contributions are as follows :
\begin{itemize}
    \item 
    We train and release the first large-scale pre-trained language model for French tweets : BERTweetFR. We make it publicly available in the transformers library and hope that it can facilitate and promote future research in analytic tasks for French tweets. 
    
    \item 
    We create and annotate the first dataset for offensiveness identification in French tweets. Such datasets already exist in several other languages and our effort fills in the gap for the French language.
    
    \item
    We create a framework and baseline for evaluating language models for French tweets. Datasets for Twitter tasks in French is very scarce and no previous work has ever combined different analytic tasks together in a unified framework.
    
\end{itemize}

With around $70\%$ of all Twitter posts being in non-English languages, the lack of corresponding language models strongly hinders the community from exploiting the information contained in these valuable resources. For future work, we plan to train and release a series of such pre-trained language models for tweets in other low resource languages. We also call upon researchers from all over the world involved in natural language processing for social media to adapt language models in their respective languages and make them publicly available.

\section*{Acknowledgements}
This research is supported by the French National research agency (ANR) via the ANR XCOVIF (AAP RA-COVID-19 V6) project. We would also like to thank the National Center for Scientific Research (CNRS) for giving us access to their Jean Zay supercomputer.

\bibliography{anthology,custom}
\bibliographystyle{acl_natbib}

\end{document}